\documentclass[acmtog]{acmart}

\usepackage{enumitem}
\copyrightyear{2025}
\acmYear{2025}
\setcopyright{cc}
\setcctype{by-nc}
\acmConference[SIGGRAPH Conference Papers '25]{Special Interest Group on Computer Graphics and Interactive Techniques Conference Conference Papers}{August 10--14, 2025}{Vancouver, BC, Canada}
\acmBooktitle{Special Interest Group on Computer Graphics and Interactive Techniques Conference Conference Papers (SIGGRAPH Conference Papers '25), August 10--14, 2025, Vancouver, BC, Canada}
\acmDOI{10.1145/3721238.3730738}
\acmISBN{979-8-4007-1540-2/2025/08}

\begin{document}

\title[ForceGrip]{ForceGrip: Reference-Free Curriculum Learning for Realistic Grip Force Control in VR Hand Manipulation}

\author{DongHeun Han}
\orcid{0000-0001-7693-7674}
\affiliation{%
  \institution{IIIXR Lab, Kyung Hee University}
  \country{South Korea}
}
\email{hand32@khu.ac.kr}

\author{Byungmin Kim}
\orcid{0009-0001-7161-6588}
\affiliation{%
  \institution{IIIXR Lab, Korea University}
  \country{South Korea}
}
\email{kbmstar1@korea.ac.kr}

\author{RoUn Lee}
\orcid{0000-0003-1596-5115}
\affiliation{%
  \institution{IIIXR Lab, Kyung Hee University}
  \country{South Korea}
}
\email{dlfhdns@khu.ac.kr}

\author{KyeongMin Kim}
\orcid{0000-0001-5168-9563}
\affiliation{%
  \institution{IIIXR Lab, Korea University}
  \country{South Korea}
}
\email{kgm031189@korea.ac.kr}

\author{Hyoseok Hwang}
\orcid{0000-0003-3241-8455}
\affiliation{%
  \institution{AIR Lab, Kyung Hee University}
  \country{South Korea}
}
\email{hyoseok@khu.ac.kr}

\author{HyeongYeop Kang}
\authornote{Corresponding Author}
\orcid{0000-0001-5292-4342}
\affiliation{%
  \institution{IIIXR Lab, Korea University}
  \country{South Korea}
}
\email{siamiz_hkang@korea.ac.kr}

\renewcommand{\shortauthors}{DongHeun Han et al.}

\begin{abstract}
Realistic Hand manipulation is a key component of immersive virtual reality (VR), yet existing methods often rely on a kinematic approach or motion-capture datasets that omit crucial physical attributes such as contact forces and finger torques. Consequently, these approaches prioritize tight, one-size-fits-all grips rather than reflecting users’ intended force levels. 
We present \emph{ForceGrip}, a deep learning agent that synthesizes realistic hand manipulation motions, faithfully reflecting the user's grip force intention.
Instead of mimicking predefined motion datasets, ForceGrip uses generated training scenarios—randomizing object shapes, wrist movements, and trigger input flows—to challenge the agent with a broad spectrum of physical interactions.
To effectively learn from these complex tasks, we employ a three-phase curriculum learning framework comprising \textit{Finger Positioning}, \textit{Intention Adaptation}, and \textit{Dynamic Stabilization}. This progressive strategy ensures stable hand-object contact, adaptive force control based on user inputs, and robust handling under dynamic conditions. Additionally, a proximity reward function enhances natural finger motions and accelerates training convergence.
Quantitative and qualitative evaluations reveal ForceGrip's superior force controllability and plausibility compared to state-of-the-art methods. 
Demo videos are available as supplementary material and the code is provided at \url{https://han-dongheun.github.io/ForceGrip}.
\end{abstract}




\begin{teaserfigure}
    \centering
    \includegraphics[width=1\linewidth]{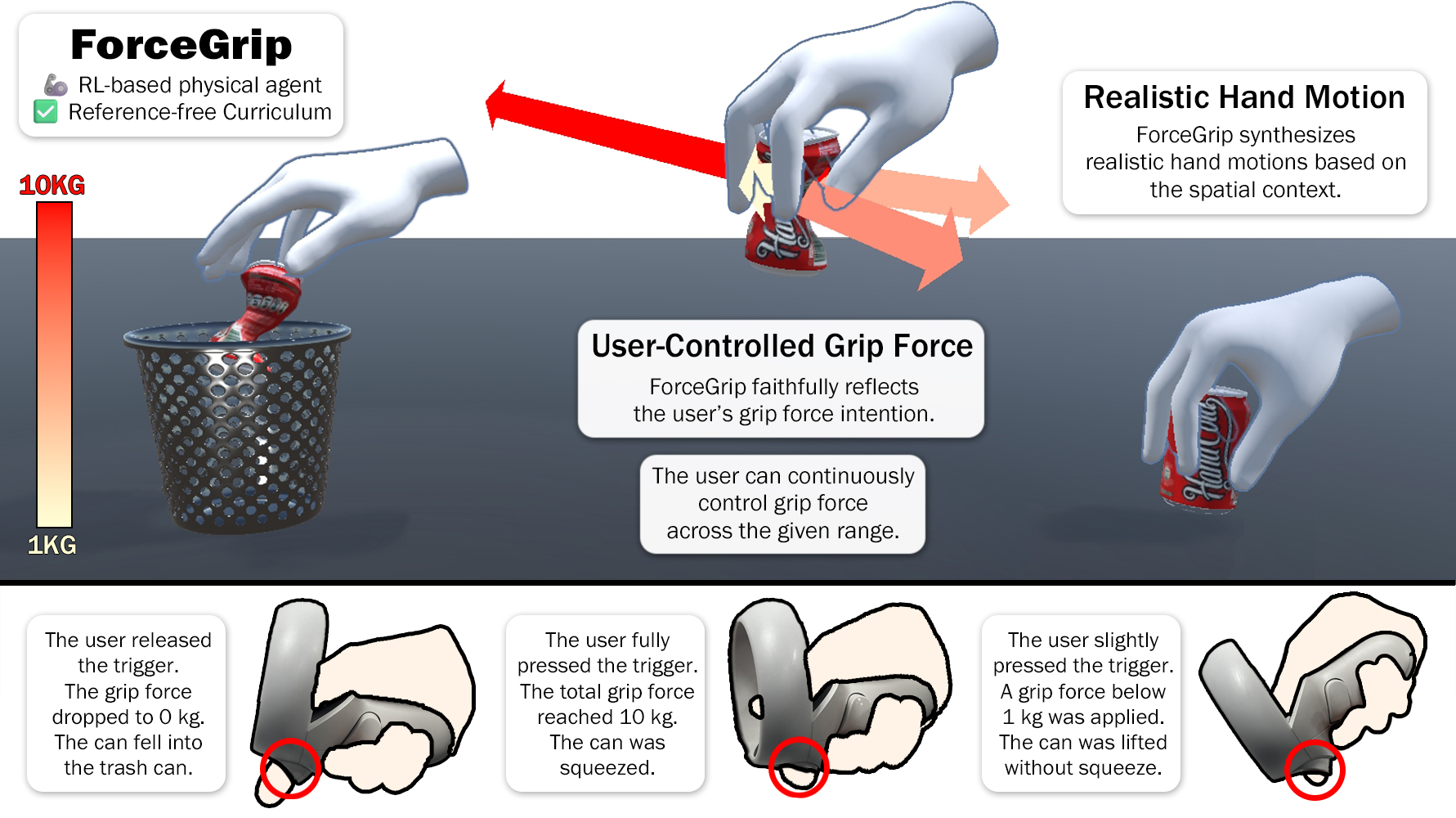}
    \caption{ForceGrip is a deep learning agent that synthesizes realistic hand manipulation motions and faithfully reflects the user's intended grip force.}
    \Description{A visual summary of the ForceGrip agent showing a hand manipulating an object, representing the agent's ability to reflect user grip force.}
    \vspace{-0.0cm}
    \label{fig:main_figure}
\end{teaserfigure}

\maketitle
\section{Introduction}
Immersive hand interaction is vital for compelling virtual reality (VR) experiences, enabling users to naturally grasp and manipulate objects.
Although controller-based interfaces —without relying on hand tracking— have become prevalent due to their accessibility, usability and responsiveness~\cite{han2023vr}, most existing solutions overlook the nuanced control of grip force. Instead, they emphasize visually plausible hand motions that rely on simple, binary trigger inputs.
This gap undermines a key dimension of realistic interaction: in physical environments~\cite{handphysicslab}, users continually fine-tune their grip force to accommodate differences in object weight, fragility, and friction, allowing objects to slip under varying force levels or remain firmly secured when needed.

Translating these nuances into VR poses considerable challenges, as conventional datasets and motion-capture techniques~\cite{kim2024damo} rarely include detailed physical attributes such as contact forces or finger torques. Consequently, it is crucial to provide diverse scenarios that involve various physical demands and ensure that these complex scenarios can be effectively learned.

To address this gap, we introduce \emph{ForceGrip}, a deep learning agent that translates VR controller inputs into realistic grip force dynamics. Rather than imitating static motion captures, ForceGrip learns from randomly generated scenarios that vary object shape, trigger signals, wrist movements and external conditions. Our curriculum learning framework systematically increases complexity through three phases: \textit{Finger Positioning}, \textit{Intention Adaptation}, and \textit{Dynamic Stabilization}. 
We further incorporate a proximity-based reward to guide natural fingertip contact, accelerating training and foster natural finger movements without imitating reference motion data.

In summary, our key contributions are as follows:
\begin{enumerate}
\item \textbf{Reference-free training.} ForceGrip synthesizes visually and physically realistic hand manipulations without relying on motion-capture data.
\item \textbf{User-controlled grip force.} By faithfully translating users’ grip force intentions into realistic hand motions, ForceGrip significantly improves interaction precision and immersion.
\item \textbf{Effective curriculum learning.} We demonstrate how incremental training phases handle diverse, high-complexity VR hand interactions with robust convergence.
\end{enumerate}

\section{Related Works}
\subsection{VR Interfaces for Physical Interaction Modeling}
\label{rel:interface}
Two principal approaches exist for hand interactions in VR: \textbf{hand-tracking} and \textbf{VR controller} interfaces. Hand-tracking interfaces rely on sensors, such as VR gloves or vision-based sensors, offering high embodiment by mirroring the user’s actual hand movements~\cite{voigt2020influence}.
However, they often use penetration-based models for grip force~\cite{holl2018efficient, quan2020realistic}, which require careful per-finger calibration and can undermine usability and confidence~\cite{voigt2020influence, kangas2022trade}. 
Furthermore, task performance depends on the sensing hardware quality~\cite{viola2022small}, restricting the general user's accessibility.

In contrast, VR controller interfaces treat the controller’s pose as the wrist proxy, with finger motions either predefined or inferred. A basic approach is the attachment-based method~\cite{oprea2019visually, shi2022grasping}, which closes fingers in fixed animations and attaches objects on contact. While intuitive, it lacks motion diversity and physical realism. VR-HandNet~\cite{han2023vr} addresses this by inferring joint torques from motion-capture data, but the absence of physical attributes in the dataset limits precise grip force control.
Meanwhile, physics-aware controllers~\cite{lee2019torc, gonzalez2020asymmetry, sinclair2019capstancrunch} and avatars~\cite{tao2023embodying} have been explored to address the limitations of standard VR controllers.

\autoref{tab:vr_methods_comparison} summarizes existing methods. Building on these insights, our study proposes a controller-based, physics-based, and learning-based approach that does not rely on reference motion data. 

\setlength{\tabcolsep}{3.5pt}
\begin{table}
\caption{Structured comparison of existing VR hand manipulation methods by presenting whether they are controller-based, involve physics simulation, employ learning-based strategies, or require a reference motion dataset.}
\vspace{-0.2cm}
\begin{tabular}{ccccc}
\hline
 & \begin{tabular}[c]{@{}c@{}}Controller\\ based\end{tabular} & \begin{tabular}[c]{@{}c@{}}Physics\\ based\end{tabular} & \begin{tabular}[c]{@{}c@{}}Learning\\ based\end{tabular} & \begin{tabular}[c]{@{}c@{}} Reference\\ Free \end{tabular} \\ \hline
\begin{tabular}[c]{@{}c@{}}Penetration\\ \cite{holl2018efficient} \\ \cite{quan2020realistic}\end{tabular} & X & O & X & O \\ \hline
\begin{tabular}[c]{@{}c@{}}Attachment \\ \cite{oprea2019visually} \end{tabular}& O & X & X & O \\ \hline
\begin{tabular}[c]{@{}c@{}}VR-HandNet \\ \cite{han2023vr}\end{tabular} & O & O & O & X \\ \hline
\textbf{\begin{tabular}[c]{@{}l@{}}ForceGrip\end{tabular}} & O & O & O & O \\ \hline
\end{tabular}
\label{tab:vr_methods_comparison}
\Description{A comparison table listing different VR hand manipulation methods. Rows represent different methods: Penetration, Attachment, VR-HandNet, and ForceGrip. Columns indicate whether each method is controller-based, physics-based, learning-based, and reference-free. Checkmarks (O) and crosses (X) show support for each feature. ForceGrip supports all four.}
\vspace{-0.4cm}
\end{table}
\setlength{\tabcolsep}{6pt}

\subsection{Physics-based Animation Training}
Physics-based animation calculates joint torques via a physics engine to account for collisions, balance, and other constraints, enabling systems to adapt dynamically to environmental changes and external forces.
Popular engines~\cite{todorov2012mujoco, coumans2016pybullet, liang2018rllib, makoviychuk2021isaac} often lack differentiability or smooth integration with deep learning. Although differentiable engines exist~\cite{hu2019difftaichi, freeman2021brax}, they see limited use due to computational overhead and restricted physics features. 
Consequently, reinforcement learning (RL)~\cite{sutton1999reinforcement} serves as the dominant method, as it trains agents via reward signals without requiring differentiability.

In general, physics-based animation research addresses three key goals: animation reproduction, task completion, and user control.

In animation reproduction research, the objective is to replicate target motions within a physically simulated environment for enhanced realism.
Techniques such as imitation reward~\cite{peng2017deeploco, peng2018deepmimic, park2019learning, won2019learning, lee2021learning1} and generative adversarial imitation learning (GAIL)~\cite{ho2016generative} are often adopted to align simulated poses with reference motion data~\cite{luo2020carl, peng2021amp, peng2022ase, hassan2023synthesizing}.
This approach preserves natural motion quality but can be limited by the coverage of available datasets.

In task completion research, agents learn specific tasks such as locomotion~\cite{heess2017emergence, klipfel2023learning}, object manipulation~\cite{kumar2015mujoco, andrychowicz2020learning, christen2022d, zhang2024artigrasp}, or specialized actions like door opening~\cite{rajeswaran2017learning} or chopstick handling~\cite{yang2022learning}.

In user control research, real-time user inputs dynamically reshape the agent’s objectives, integrating animation reproduction for smooth, natural responses.
Examples include determining walking directions~\cite{bergamin2019drecon, wang2020unicon, won2022physics}, generating diverse skill animations~\cite{lee2021learning2}, and controlling animations through natural language~\cite{juravsky2022padl, cui2024anyskill}. 
For hand-specific control, VR controllers can provide user signals used for finger motion simulation~\cite{han2023vr}.

Building on these foundations, our research focuses on user control with an emphasis on precise grip force translation.

\subsection{Curriculum Learning}
Curriculum learning~\cite{bengio2009curriculum} organizes training tasks in increasing order of difficulty, a strategy shown to accelerate convergence and improve performance in deep learning~\cite{weinshall2018curriculum, baker2019emergent, hacohen2019power}. 

Within physics-based animation, this approach helps agents learn complex skills more efficiently.
For example, \cite{won2021control} first trained basic skills and then introducing adversarial environments. Similarly, \cite{liu2022motor, luo2023perpetual, zhang2023learning, haarnoja2024learning} guide agents to learn increasingly diverse and challenging skills. 
Although automated task sequence generation has been explored~\cite{graves2017automated, florensa2018automatic, silva2018object, narvekar2018learning}, no universal standard exists for defining task difficulty~\cite{narvekar2020curriculum}.
Instead, manual structuring remains highly effective~\cite{justesen2018illuminating, cobbe2019quantifying, wang2019paired}.

In this work, we devise a structured curriculum to address the challenges of grip force control in VR. 

\section{Overview}
\autoref{fig:main_figure} depicts our training pipeline. 
We employ Unity with NVIDIA PhysX for real-time physics, updating at 120 Hz while the learning agent acts at 30 Hz. Each time step~$t$, the agent outputs torques for finger joints, which a proportional-derivative (PD) controller~\cite{tan2011stable} translates into hand poses and drives object interactions at time~$t+1$. 
During training, the wrist follows pre-set motion paths, whereas in actual usage it follows VR controller movements.

Instead of relying on motion-capture data, we generate diverse scenarios encompassing varied object shapes, user inputs, and wrist motions.
However, this variability increases learning complexity.
To address this, our curriculum strategy partitions training into simpler tasks, followed by progressively more dynamic conditions.

To train the agent, we leverage RL.
We use proximal policy optimization (PPO)~\cite{schulman2017proximal}, running 576 concurrent agents with different seeds.
Detailed hyperparameter settings are given in the supplementary materials.
We employ \textit{early termination}~\cite{peng2018deepmimic} to stop an episode if an object moves more than 10 cm away from the wrist, avoiding wasted computation on irrecoverable states.
This mechanism increases efficiency and prevents spurious rewards, ultimately leading to a more stable policy.

\begin{figure}
    \centering
    \includegraphics[width=1\linewidth]{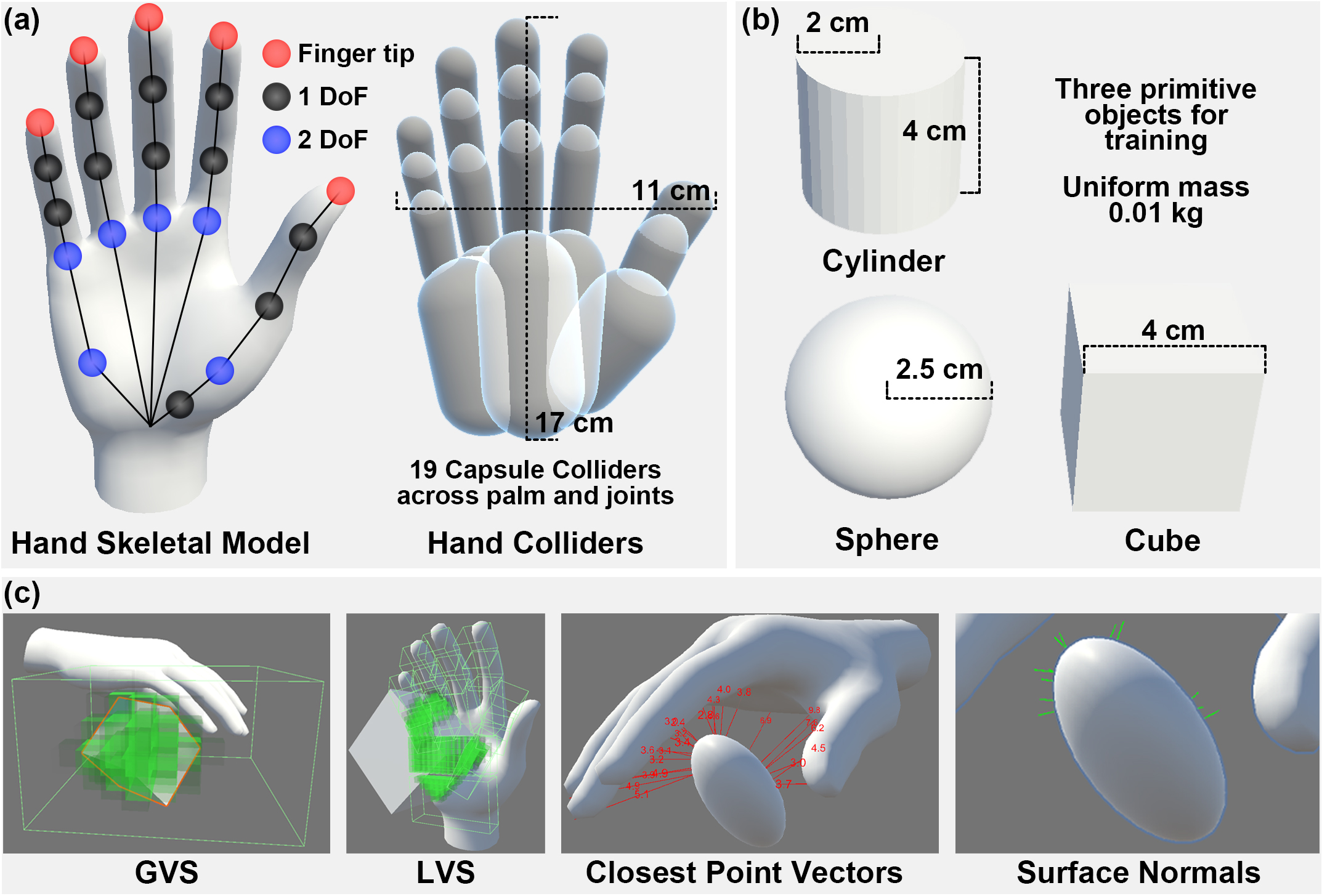}
    \vspace{-0.6cm}
    \caption{(a) Hand model from Meta XR Core SDK. (b) Objects: sphere, cube, and cylinder. (c) Sensors: GVS, LVS, closest and normal vector.}
    \Description{Illustration showing (a) the hand model with joint structures and colliders, (b) the three basic object types used for training—sphere, cube, and cylinder, and (c) the sensor setup including global voxel sensors (GVS), local voxel sensors (LVS), and closest point with normal vector visualization.}
    \label{fig:handmodel_objects_sensors}
    \vspace{-0.4cm}
\end{figure}

\subsection{Hand Model and Objects}
We utilize the Meta XR Core SDK hand model, which has 17 finger joints (23 total DoFs) and 19 non-self-colliding capsule colliders spread across the fingers and palm (\autoref{fig:handmodel_objects_sensors}(a)).
To learn various grip conditions, we use three primitive object shapes—sphere, cube, and cylinder~(\autoref{fig:handmodel_objects_sensors}(b))—each with a uniform mass of 0.01 kg. 
Since dropping an object results in early termination, the trained agent tends to apply excessive force to avoid dropping, even when this force does not match the target force. To prevent this mismatch, we mitigate the effect of mass by using lightweight objects.

\begin{figure*}[t!]
    \centering
    \includegraphics[width=1\linewidth]{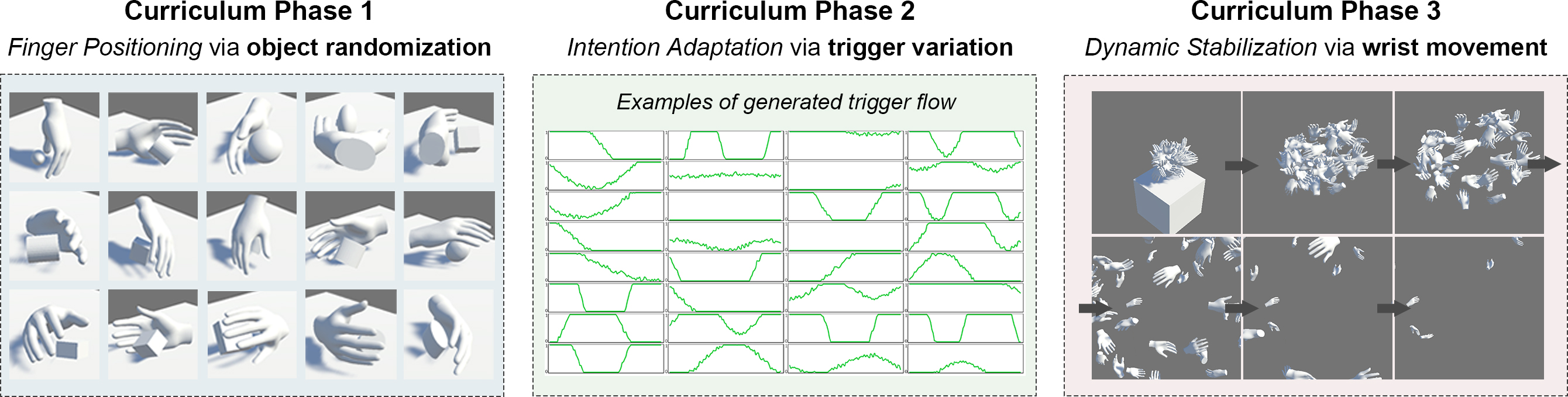}
    \vspace{-0.6cm}
    \caption{Overview of our training pipeline. Rather than relying on motion-capture data, we generate diverse scenarios of varying object shapes, user inputs, and wrist motions; a three-phase curriculum progressively increases task complexity.}
    \Description{Curriculum learning framework diagram with three sequential stages: Finger Positioning, Intention Adaptation, and Dynamic Stabilization. Each stage introduces increasing complexity to train grip force control.}
    \label{fig:curriculum_learning}
    \vspace{-0.3cm}
\end{figure*}

\subsection{State and Action}
ForceGrip is a deep RL agent that receives an environment state $s \in \mathbb{R}^{3023}$ and outputs an action $a \in \mathbb{R}^{23}$ at each time step. The state $s$ comprises three primary components: hand model information ($H$), object information ($O$), and task information ($T$).

Hand model information $H = \{p, q, v, w, \alpha\}$ includes joint positions $p \in \mathbb{R}^{66}$, joint DoF angles $q \in \mathbb{R}^{23}$, joint linear velocities $v \in \mathbb{R}^{51}$, joint DoF velocities $w \in \mathbb{R}^{23}$, and joint DoF accelerations $\alpha \in \mathbb{R}^{23}$. 

Object information $O = \{o_g, v_{gvs}, v_{lvs}, c, n\}$ includes the gravity vector $o_g \in \mathbb{R}^3$ in wrist coordinates, a global voxel sensor $v_{gvs} \in \mathbb{R}^{450}$ attached to the wrist, employing 2 cm cells, local voxel sensors $v_{lvs} \in \mathbb{R}^{2160}$ attached to each collider, employing 0.5 cm cells, closest point vectors $c \in \mathbb{R}^{69}$ from relevant joints to the object's surface, and corresponding surface normals $n \in \mathbb{R}^{69}$.
Inspired by previous work~\cite{zhang2021manipnet, han2023vr}, this dual-resolution sensing strategy provides a balance between detail and generality, minimizing overfitting to a specific object shape. Object information is illustrated in~\autoref{fig:handmodel_objects_sensors}.

Task-related information $T = \{f, u, p\}$ comprises current force vectors for each collider $f \in \mathbb{R}^{57}$, user trigger signal history $u \in \mathbb{R}^{6}$ over the past 6 frames (0.2 s), and previous action $p \in \mathbb{R}^{23}$. 
By incorporating past user inputs and recent actions, the agent can regulate grip forces more smoothly and maintain consistent hand-object contact.

\subsection{Reward function}
The agent’s primary goal is to translate user trigger signals into physically plausible grip forces.
We define a \textit{force reward} $r_f$ as follows:
\begin{equation}
    r_f = \exp[-1.0(||(\sum_{k}||f_t^k||)-f_t^{target}||^2)]
\end{equation}
where $f_t^k$ represents the per-collider force vectors at time $t$, and $f_t^{target}$ is the target grip force derived from the user’s input by 
\begin{equation}
    f_t^{target} = f_{max} \cdot \bar{u}_{t}
\end{equation}
Here, $f_{max} = 10~\mathrm{kgf}$ is the maximum grip force, and $\bar{u}_{t}$ is the average trigger signal over recent six frames. 
Note that the value of $f_{max}$ can be adjusted for different applications to accommodate varying device or interaction requirements.

Relying solely on a task-centric reward can yield stiff, unnatural hand motions, so we introduce a \textit{proximity reward} $r_p$ to guide initial contact:
\begin{equation}
    r_{p} = \sum_j{w^j\exp[-0.07||c_t^j||^2]}
\end{equation}
where $c_t^j$ is the distance vector from the $j$-th joint to the object surface, and $w^j$ is a weight emphasizing fingertip-first contact. We assign higher weights (0.0625) to the five end-effectors and their associated joints, and lower weights (0.03125) elsewhere. Specifically, joints or end-effectors within \textasciitilde1cm of an object earn reward >0.9 before applying the weighted sum, encouraging natural contact.

The final reward combines these two terms:
\begin{equation}
    r = r_f + r_{p}
\end{equation}

Rather than ensuring all objects can always be lifted, our objective is to solely reflect user-intended force; insufficient force causes objects to slip. The user controls only the amount of force, not whether to grasp or release, making such slips a natural outcome by design.

\section{Training}
\subsection{Training scenario generation}
To ensure broad adaptability and avoid limited-coverage datasets, we generate 3-second (90-frame) training scenarios rather than relying on motion-capture data. 
Although many VR tasks lack a strict endpoint, we treat these scenarios as repeating episodes in real-world deployment. 

Our training scenarios incorporate three key factors:
\paragraph{Object randomization} Each episode randomly scales objects along $x$, $y$, and $z$-axis by factors between 0.5 and 1.5 and arbitrarily positions them within a predefined volume in front of the hand.
During the first 0.5 seconds, object stabilization is performed. Each object is rendered kinematic (i.e., not influenced by physics) to stabilize poses, allowing the agent’s fingers to approach the object without premature slipping. 

\paragraph{Trigger Variation.} We synthesize noisy trigger signals by combining sinusoidal functions with Gaussian noise. Varying frequencies, amplitudes, and offsets prevents overfitting to any single control pattern. Supplementary materials detail the parameters and sample waveforms.

\paragraph{Wrist movement} 
After object stabilization, we remove the floor and introduce randomized wrist (up to $2~m/s^2$) and angular (up to $360^\circ/s^2$) accelerations at each physics simulation step. 
This reflects the dynamic real-world conditions and challenges the agent to maintain stable grip forces.

\subsection{Curriculum learning}
Because these random, high-complexity scenarios pose significant training challenges, we devise a three-phase curriculum (\autoref{fig:curriculum_learning}): \textit{Finger Positioning}, \textit{Intention Adaptation}, and \textit{Dynamic Stabilization}. Each phase builds on the skills developed in the preceding phase, progressively training the agent to handle increasingly dynamic and realistic conditions.

\subsubsection{Phase 1: Finger Positioning}
This phase trains the agent to establish accurate finger contact.
To minimize distractions,  only \textbf{object randomization} is active; \textit{trigger variation} and \textit{wrist movement} are disabled. The wrist remains fixed with its pose randomly initialized, and user triggers are set to constant random values within [0,1]. This setup accelerates the agent’s mastery of foundational behaviors such as collision handling and initial grip formation.

\subsubsection{Phase 2: Intention Adaptation}
Building on Phase~1, we add \textbf{trigger variation}, requiring the agent to interpret fluctuating signals and apply proportionate forces in real time.

\subsubsection{Phase 3: Dynamic Stabilization}
Building on Phase~2, we add \textbf{wrist movement}, introducing abrupt orientation and acceleration changes. 
Successful policies must preserve contact forces without losing grip, even when gravity and inertial forces vary substantially.

In our experiments, phase 1 trained for 1200 epochs, followed by 400 epochs of Phase 2, and then Phase 3 until reward convergence (800 epoches empirically).

\section{Curriculum and Ablation Study}
\label{eval}
We evaluated our curriculum learning strategy and conducted ablation studies to assess the impact of specific design choices. 

\subsection{Test Conditions}
Recall that our curriculumn comprises three sequential phases: \textit{Finger Positioning} (P1), \textit{Intention Adaptation} (P2), and \textit{Dynamic Stabilization} (P3). 
This aims to decompose randomized factors—such as varying trigger inputs and wrist motions—into progressively challenging phases.

To analyze alternative curriculum design, we introduce two additional phase variants: \textit{Intention Adaptation} but with wrist movement instead of trigger variation (P2-b) and \textit{Dynamic Stabilization} but with trigger variation instead of wrist movement (P3-b). 

Based on these phases (P1, P2, P2-b, P3, P3-b), we define six distinct curricula: 
\begin{itemize}
\item \textit{C1 (Baseline)}: P1 $\rightarrow$ P2 $\rightarrow$ P3.
\item \textit{C2 (Swapper Factor)}: P1 $\rightarrow$ P2-b $\rightarrow$ P3-b.
\item \textit{C3 (Merged P2 and P3)}: P1 $\rightarrow$ P2+P3.
\item \textit{C4 (Merged P1 and P2)}: P1+P2 $\rightarrow$ P3.
\item \textit{C5 (Merged P1 and P2 for C2)}: P1+P2-b $\rightarrow$ P3-b.
\item \textit{C6 (No Curriculum)}: P1+P2+P3.
\end{itemize}

Here, ``merged" phases indicate that two sets of randomization factors (e.g., trigger variation and wrist movement) are introduced simultaneously rather than sequentially.

In addition to examining the curriculum, we conduct two additional ablation experiments to investigate the impact of design choices: 
\begin{itemize}
\item \textit{PR $\times$}: Remove the \textit{proximity reward}, originally included to encourage lifelike fingertip contact and natural hand poses.
\item \textit{EL $\times$}: Remove the \textit{encoder layers} responsible for extracting domain-relevant information from each feature subset. 
\end{itemize}

Both ablation conditions use the C1 curriculum with identical phase-transition epochs to ensure fair comparison. Preliminary analysis confirmed these transitions did not hinder convergence. For \textit{EL $\times$}, voxel-based encoders for $v_{gvs}$ and $v_{lvs}$ were retained to maintain trainability due to their large feature sizes.

\begin{figure}
    \centering
    \includegraphics[width=1\linewidth]{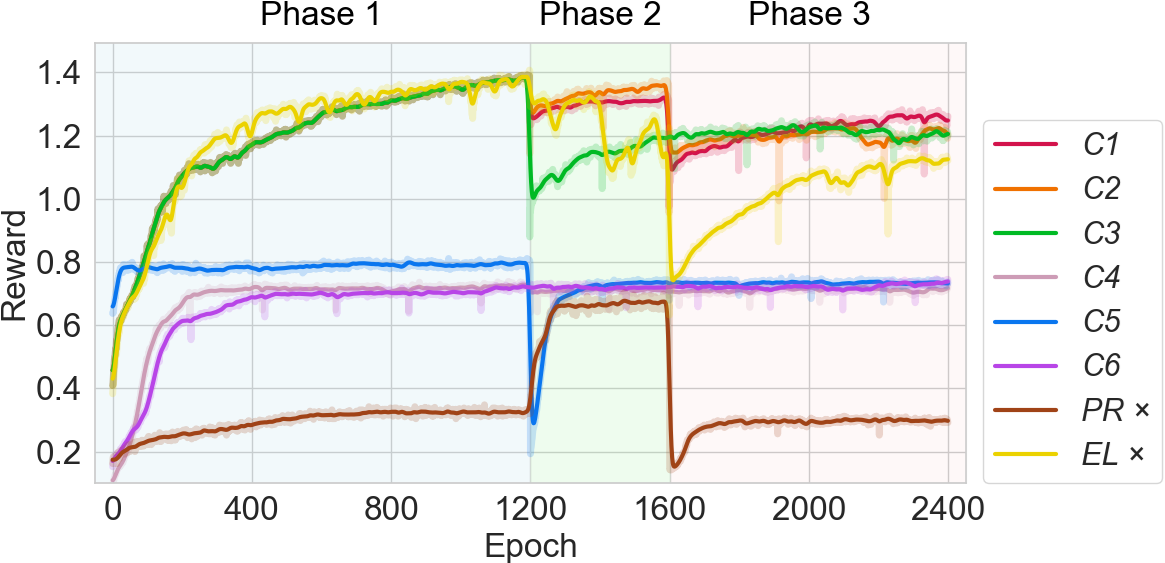}
    \vspace{-0.7cm}
    \caption{Reward curves for the different curriculum designs (\textit{C1}–\textit{C6}) and ablation experiments (\textit{PR $\times$}, \textit{EL $\times$}). The vertical lines at epochs 1200 and 1600 mark the transitions between Phases 1, 2, and 3. Note that in the \textit{PR $\times$}, the curve is doubled to match scale, as only a single reward term remains.}
    \Description{Line graph comparing training rewards over epochs for six curriculum learning strategies and two ablation conditions. The x-axis shows training epochs, and the y-axis shows normalized reward values. Vertical lines indicate transitions between training phases. ForceGrip with standard curriculum shows the highest convergence.}
    \label{fig:curriculum_reward_result}
\vspace{-0.4cm}
\end{figure}

\subsection{Evaluation Results}

To investigate the training progression, we measured reward in every epoch for each curriculum and ablation setting (\autoref{fig:curriculum_reward_result}). Curricula that do not begin with P1 (\textit{C4}, \textit{C5}, and \textit{C6}) exhibit noticeably lower rewards throughout training. This outcome emphasizes learning stable finger contact before introducing additional complexities. Among the curricula, our proposed approach (\textit{C1}) achieves the highest overall reward.

\renewcommand{\arraystretch}{1.1}
\setlength{\tabcolsep}{3.5pt}
\begin{table}[h]
\vspace{-0.2cm}
\caption{Quantitative evaluation results. The highest score in each row is \underline{\textbf{bold and underlined}}, and the second-highest is \underline{underlined} only.}
\vspace{-0.2cm}
\begin{tabular}{lcccccccl}
\hline
& \textit{C1} & \textit{C2} & \textit{C3} & \textit{C4} & \textit{C5} & \textit{C6} & \textit{PR $\times$} & \textit{EL $\times$} \\ \hline
ESR (\%) & \underline{81.30} & 73.96 & 74.33 & 75.33 & 77.06 & 79.38 & 43.26 & \underline{\textbf{83.44}} \\ \hline
PR       & \underline{0.635} & 0.601 & 0.595 & 0.589 & 0.605 & 0.607 & 0.014 & \underline{\textbf{0.657}} \\ \hline
FR       & \underline{\textbf{0.607}} & 0.594 & \underline{0.601} & 0.123 & 0.120 & 0.126 & 0.133 & 0.460 \\ \hline
\end{tabular}
\label{tab:quantitative_result}
\Description{A table showing quantitative evaluation results across six curriculum variants and two ablation studies. Metrics include Episode Success Ratio (ESR), Proximity Reward (PR), and Force Reward (FR). Bold and underlined values indicate the best performance, and underlined values indicate the second best. C1 performs best in force control.}
\vspace{-0.2cm}
\end{table}
\setlength{\tabcolsep}{6pt}
\renewcommand{\arraystretch}{1}
For a more in-depth analysis, we collected data from 100,000 episodes and measured three key metrics (\autoref{tab:quantitative_result}): episode success ratio (ESR), average proximity reward (PR), and average force reward (FR).
While \textit{C4}, \textit{C5}, and \textit{C6} maintain relatively acceptable ESR and PR scores, their lower FR scores reveal difficulties in managing force control when advanced randomization (e.g., full wrist movement or dynamic triggers) is introduced early. 
By contrast, \textit{C2} and \textit{C3}—which incorporate key phases in a more structured sequence—yield moderate ESR but fall below \textit{C1} in overall stability.
Removing the proximity reward (\textit{PR $\times$}) significantly reduces performance, particularly in ESR and PR, but also affects FR. 
Although eliminating encoder layers (\textit{EL $\times$}) achieves a higher ESR and PR, it simultaneously lowers the FR score. This outcome suggests that the network, lacking encoder-based feature extraction, struggles to incorporate critical low-dimensional inputs (e.g., $u \in \mathbb{R}^6$), causing the agent to adopt an overly forceful grip rather than refining grip force precision.

Overall, the results confirm that our structured curriculum improves both success rates and grip realism. Ablation studies highlight the impact of design choices such as the proximity reward and encoder layers on precise grip force control.

\section{User Study}
To present the superiority of our method, \textit{ForceGrip} (FG), we conducted a comparative user study against two state-of-the-art VR controller interface methods: \textit{Attachment} (AT)~\cite{oprea2019visually} and \textit{VR-HandNet} (VH) ~\cite{han2023vr}.

\subsection{Evaluation Metrics}
To evaluate visual realism, we adopted the Realism Questionnaire (RQ) proposed by \cite{oprea2019visually} containing 14 items across three categories: motor control, finger movement realism, and interaction realism. The individual scores are then aggregated into category-specific and overall indices, providing insights into perceptual realism from a purely visual standpoint.

To evaluate physical user experience, we devised a \textit{Force Realism Questionnaire} (FRQ). Inspired by~\cite{gonzalez2018avatar} and tailored to physically simulated hand interactions, the FRQ retains the RQ’s three-category framework but focuses on evaluating \emph{grip force} accuracy and user intention. Detailed FRQ items and scoring procedures are provided in the supplementary material.

To ensure that user comfort did not skew any evaluations, we employed the Simulator Sickness Questionnaire (SSQ)~\cite{kennedy1993simulator} at the beginning and conclusion of each condition. 

We applied the Friedman test to detect statistically significant differences, as the data did not meet the normality assumptions required by parametric tests. For post-hoc analysis, we conducted Wilcoxon signed-rank tests. Detailed statistical comparison results are provided in the supplementary materials.

\subsection{Participants and Apparatus}
We recruited 20 participants (13 males and 7 females; $\mu=23.1, \sigma=1.55$), comprising 11 individuals with prior VR experience and 9 newcomers. 
Each participant practiced all three interface methods (FG, VH, AT) for 5 minutes each, with additional practice time available upon request to ensure comfort with the controls.

We used a Meta Quest 2 headset and stock controller, reflecting a typical consumer-grade VR setup, powered by a desktop featuring an NVIDIA RTX 3080 GPU and an AMD Ryzen 5 5600G CPU. This setup provided consistent performance across all trials, minimizing hardware-induced variability.

\subsection{E1: Pick-and-place --- Method}
\begin{figure*}[h!]
    \centering
    \includegraphics[width=0.99\linewidth]{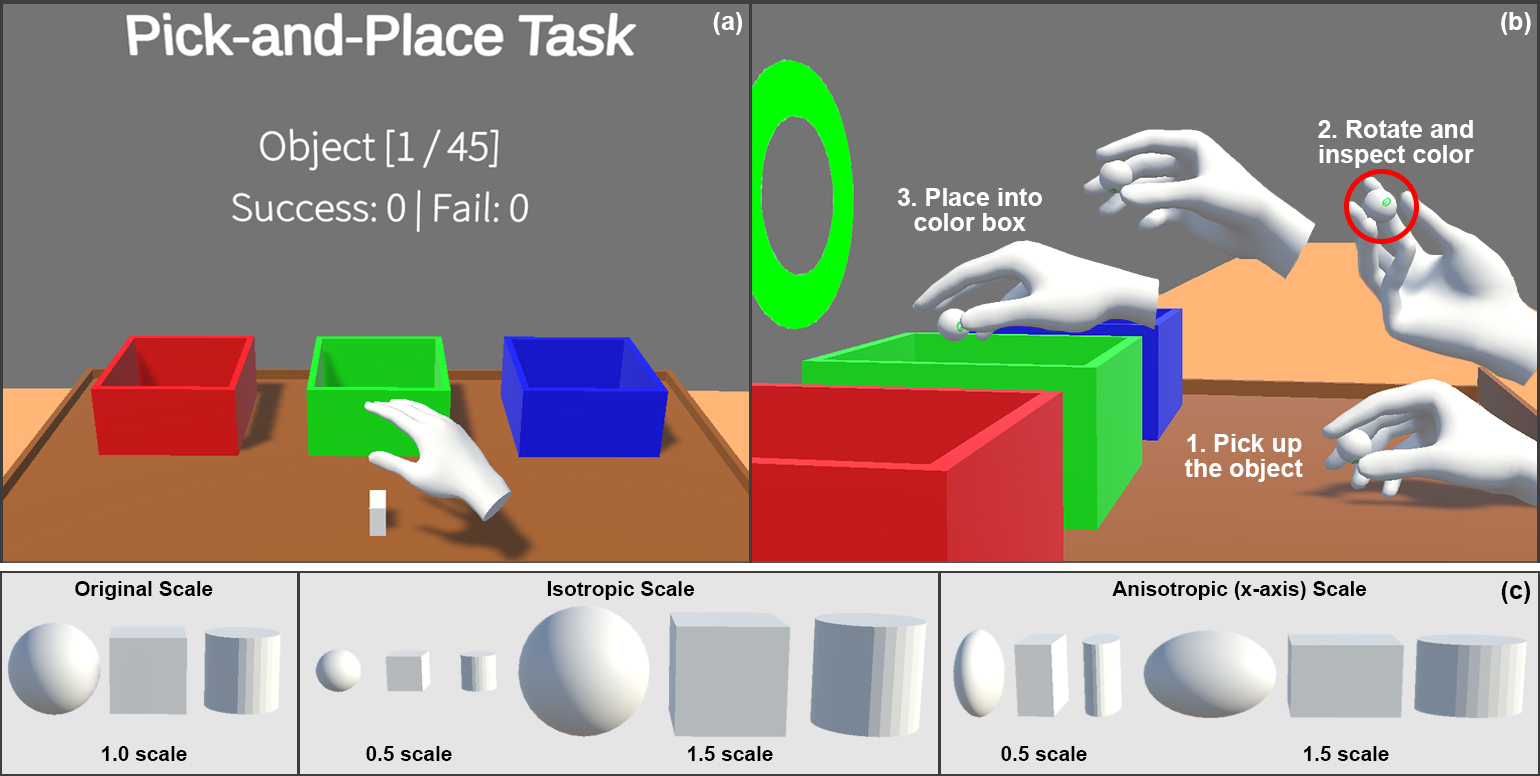}
    \vspace{-0.4cm}
    \caption{\textbf{Pick-and-place task}. (a) Randomly generated objects with color label pairs (R, G, or B). (b) Procedure: pick up the object, rotate and inspect its color label, and place it into the corresponding color box. (c) 15 candidate objects are displayed at three scales: original, isotropic, and anisotropic.}
    \Description{Image illustrating the pick-and-place task. Includes a set of virtual objects with colored labels, the user inspecting the label, and placing the object into a matching colored box. Shows three object types in original, isotropic, and anisotropic scales.}
    \label{fig:pick_and_place}
    \vspace{-0.0cm}
\end{figure*}

To evaluate visual realism in common object-handling scenarios, such as grasping, rotating, and translating, we designed a \textit{Pick-and-Place} task~(\autoref{fig:pick_and_place}). 

Each participant completed three blocks in random order, with each block dedicated to one of three methods: FG, VH, and AT. Within each block, participants encountered 15 distinct objects—three at their original scale, six isotropically scaled along either $x$-, $y$-, or $z$-axis (by 0.5 and 1.5), and six anisotropically scaled along the $x$-axis (also by 0.5 and 1.5). Each object was assigned one of three color labels (R, G, or B), resulting in 45 unique object–label pairs per block.

With these 45 object-label pairs, participants were asked to perform \textit{Pick-and-Place} task. Whenever a random object appeared in front of them, participants were instructed to pick it up, inspect the color label (R, G, or B) on its underside, and place it into the corresponding colored box.
These 45 pairs were presented in a randomized order, ensuring no repeated pairings within the same block. Consequently, each participant completed a total of 135 \textit{Pick-and-Place} (3 blocks $\times$ 45 pairs per block).
Each block took approximately five minutes, followed by a two-minute break to mitigate fatigue.
At the end of each block, participants completed RQ.

\begin{figure}[t!]
    \centering
    \includegraphics[width=1\linewidth]{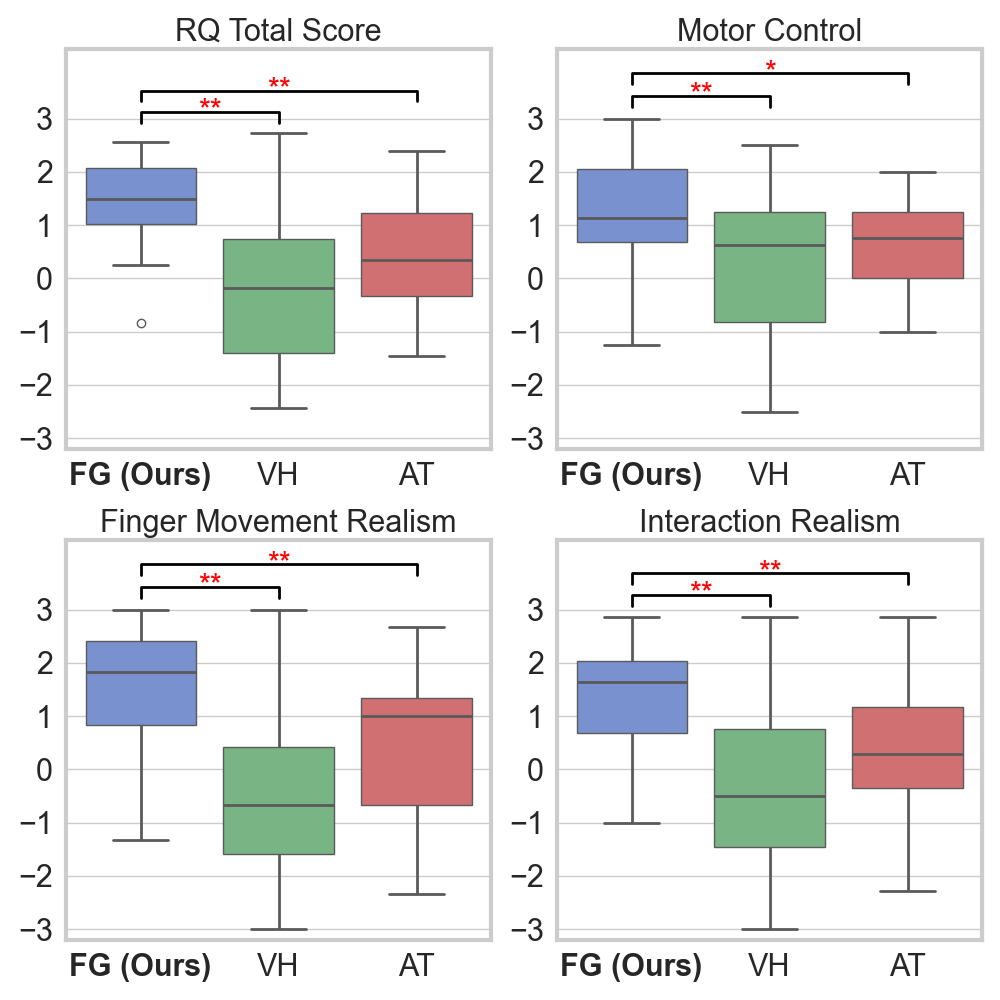}
    \vspace{-0.7cm}
    \caption{RQ scores for the Pick-and-Place task, using a 7-point Likert scale ranging from -3 (strongly disagree) to +3 (strongly agree). Brackets indicate statistical significances ($*$ : $p$ $<$ .05, $**$ : $p$ $<$ .01). }
    \Description{Box plots showing RQ scores for the Pick-and-Place task across four categories: Total Score, Motor Control, Finger Movement Realism, and Interaction Realism. Each category compares ForceGrip, VR-HandNet, and Attachment methods. ForceGrip consistently shows higher median scores. Statistical significance between methods is indicated with asterisks.}
    \label{fig:pick_and_place_result}
    \vspace{-0.3cm}
\end{figure}

\subsection{E1: Pick-and-Place --- Results}
In terms of RQ scores, a Friedman test revealed significant differences across all interfaces ($p < .01$). 
As shown in \autoref{fig:pick_and_place_result}, post-hoc Wilcoxon signed-rank tests confirmed that FG achieved significantly higher RQ scores than both VH and AT.
Additionally, there was no significant change in SSQ scores before and after the experiment, indicating minimal motion-sickness effects.

Participants provided qualitative observations that illuminated each technique’s strengths and limitations.
For FG, participants praised it’s ability to ``\textit{stably and realistically grasp small objects}” (P3, P7, P12) and ``\textit{naturally adapt finger motions to varied shapes}” (P6, P9, P11, P13, P14, P19, P20). 
Some, however, experienced occasional ``\textit{object slipping}” (P3, P6, P10), likely due to the overly rounded fingertip shape of the capsule collider model provided by the SDK.

For VH, participants praised it’s ability to ``\textit{natural dropping of objects, much like real life}” (P4, P8), highlighting the advantages of physics-based animation.
Yet, many participants criticized its ``\textit{unnatural and predefined finger-closing motion}” (P1, P2, P4, P6, P8, P12, P13, P16, P19) and ``\textit{difficulty handling smaller objects}" (P3, P7, P10, P12, P14, P18), suggesting limited adaptability—a consequence of its narrower motion-capture dataset coverage. 

For AT, Participants observed ``\textit{naturally aligned finger posture for larger objects}” (P4, P6, P7, P12, P16, P18), but found ``\textit{grasping small items challenging}” (P4, P6, P7, P11, P12, P13, P16, P18). Others noted that ``\textit{unnatural and inconvenient experience stem from actions only triggered after palm sensor activation}” (P1, P3, P8, P10, P13, P18). These reflect AT’s inability to adapt various object shapes due to the inherent limitation of the kinematic approach. 

\begin{figure*}[h!]
    \centering
    \includegraphics[width=1\linewidth]{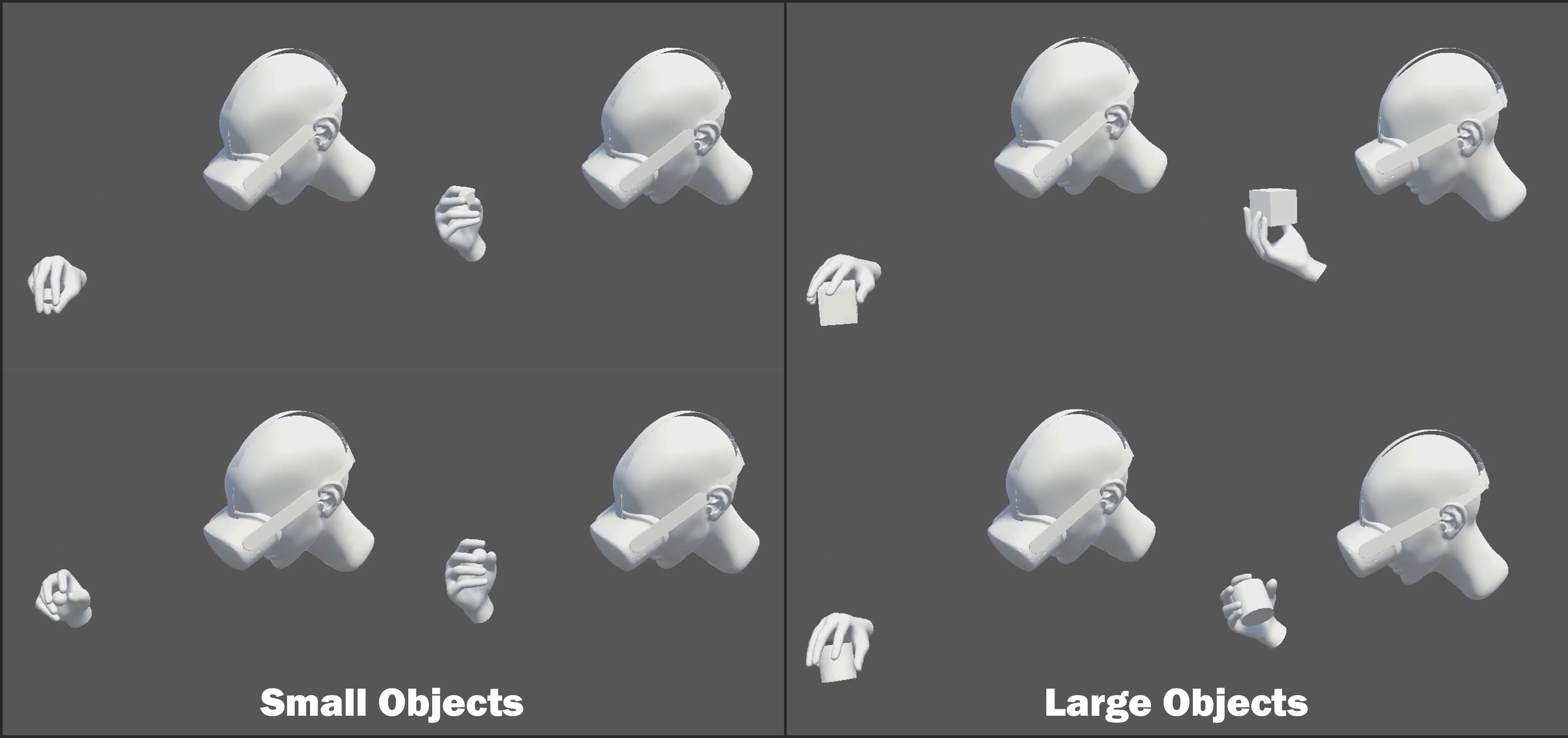}
    \vspace{-0.7cm}
    \caption{\textbf{User study interactions with small and large virtual objects.} Participants interact with virtual objects of different sizes, demonstrating ForceGrip’s adaptability across object scales. Small objects are delicately grasped using the fingertips, while large objects are securely held by widening the finger spread.}
    \Description{Photographs of participants interacting with small and large virtual objects in VR. Small objects are grasped with fingertips, while large objects require full-hand grasps. Demonstrates adaptability of hand model to object scale.}
    \label{fig:small_large_objects}
\end{figure*}

\begin{figure*}[h!]
    \centering
    \includegraphics[width=1\linewidth]{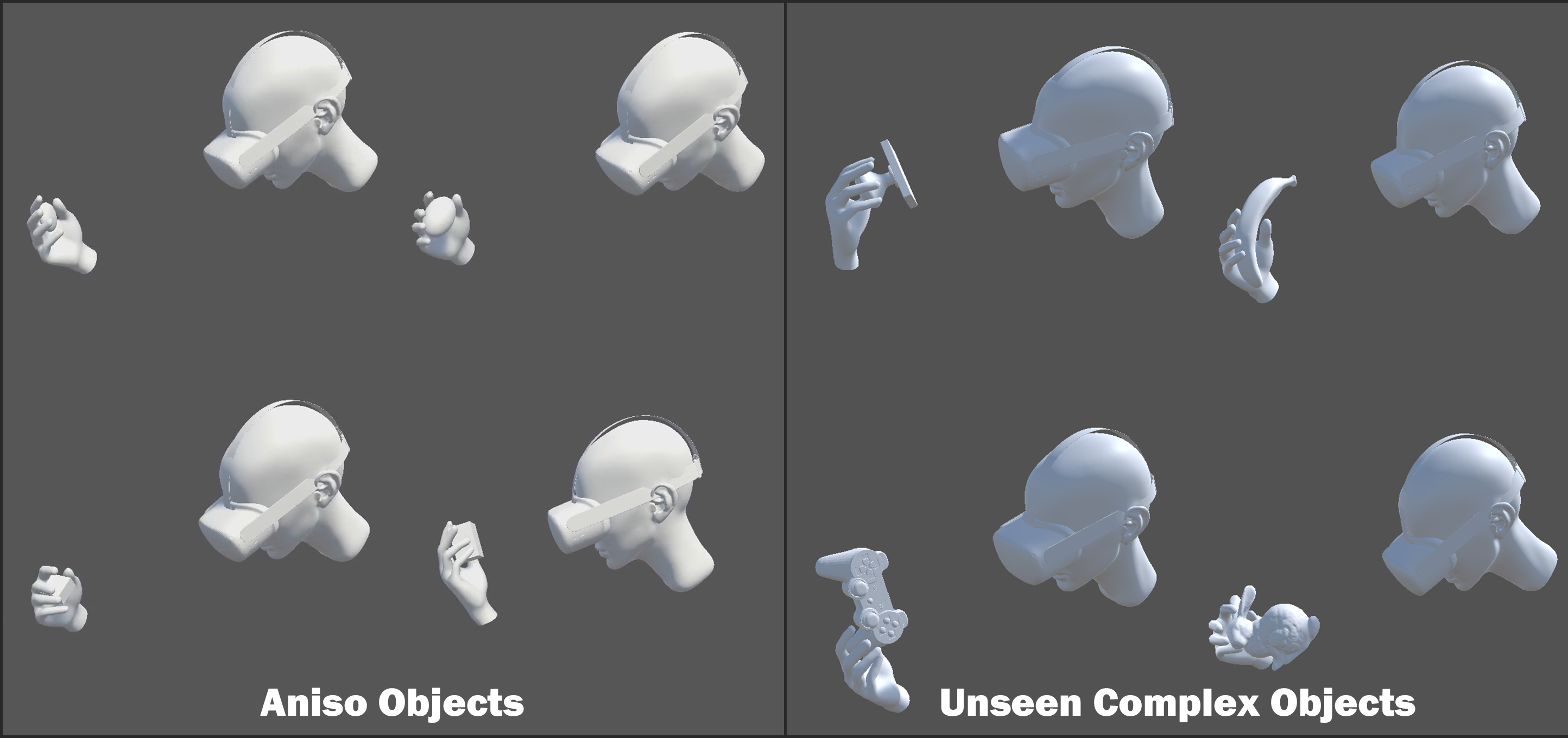}
    \vspace{-0.7cm}
    \caption{\textbf{User study interactions with anisotropically scaled virtual objects and internal test interactions with unseen complex-shaped virtual objects.} Participants interact with anisotropically scaled objects and internally tested unseen complex-shaped objects from~\cite{taheri2020grab}, demonstrating ForceGrip’s generalizability across varying object geometries. Objects with anisotropic scaling require adaptive finger positioning to maintain stable grasps, while complex-shaped objects demand fine-grained adjustments to conform to irregular surfaces.}
    \Description{Images showing participants interacting with anisotropically scaled and complex-shaped virtual objects. Finger motions adapt to object geometry to maintain realistic grips. Shows generalizability to unseen shapes.}
    \label{fig:aniso_complex_objects}
\end{figure*}

Overall, this experiment indicates that \textit{ForceGrip} provides superior visual realism for typical object manipulation tasks. Examples of user interactions with various object sizes and shapes, supporting these findings, are shown in~\autoref{fig:small_large_objects} and~\autoref{fig:aniso_complex_objects}.

\subsection{E2: Can squeeze --- Method}

\begin{figure*}[h!]
    \centering
    \includegraphics[width=0.99\linewidth]{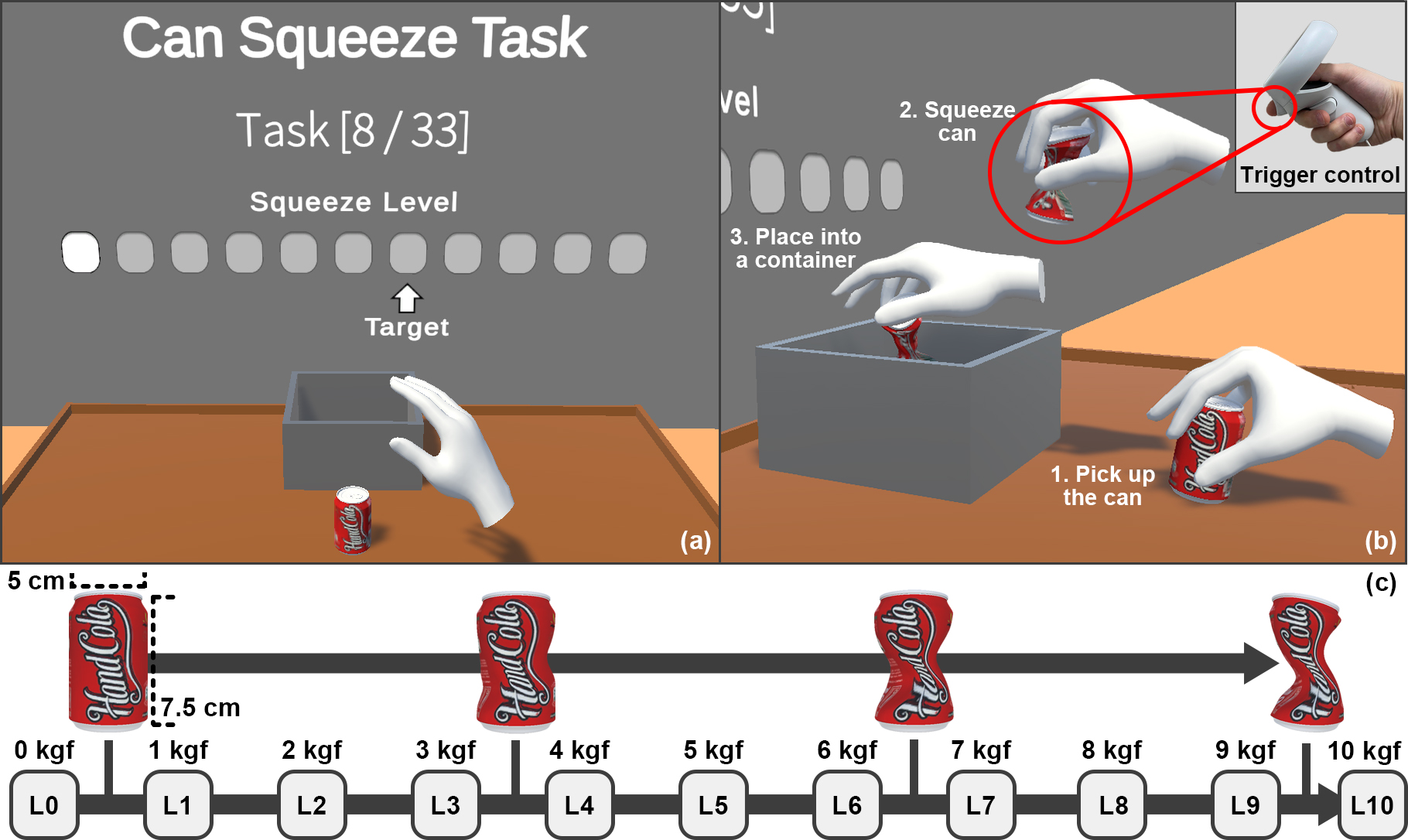}
    \vspace{-0.4cm}
    \caption{\textbf{Can squeeze task}. (a) Generated can object with a random target level. (b) Procedure: pick up the can, squeeze it via trigger control, and place it into the container. (c) Can deformation levels (L0: no deformation - L10: maximum deformation) visualized with mesh representations.}
    \Description{Images illustrating the can squeeze task. A virtual can with a target deformation level appears, the user picks it up, applies trigger pressure to deform the can, and places it into a trash can. Visuals show mesh deformation from level 0 to 10.}
    \label{fig:can_squeeze}
    \vspace{-0.1cm}
\end{figure*}

To evaluate how precisely users can translate controller-trigger inputs into physically simulated grip forces, we introduced the \textit{Can Squeeze} task (\autoref{fig:can_squeeze}). 
Because AT does not support physical interactions, it was excluded from this experiment.

Participants interacted with a virtual can modeled after a standard beverage can and were asked to deform it to a target level from 0 (none) to 10 (maximum). Each discrete level represented a 1 kgf integer increase. We used this 11-level scale to balance force granularity with the limited sensitivity of VR triggers, as finer subdivisions made it difficult for participants to modulate shallow trigger inputs in preliminary tests.

At the start of each trial, a can appeared in front of the user along with a random target deformation level on the display. The participant adjusted the VR controller’s trigger to deform the can to match the target level, then placed the can into a container. Visual and auditory feedback indicated deformation through mesh updates and sound effects.

Each participant completed two blocks, with one block dedicated to the FG and the other to VH. The order of these blocks was counterbalanced across participants. Within each block, 33 trials were conducted by repeating every target deformation level (0–10) three times, in randomized order.
Consequently, each participant performed a total of 66 trials (2 blocks $\times$ 33 trials per block). Each block took approximately five minutes to complete, followed by a two-minute break to mitigate fatigue.
At the end of each block, participants completed FRQ. 

\begin{figure}[t!]
    \centering
    \includegraphics[width=1\linewidth]{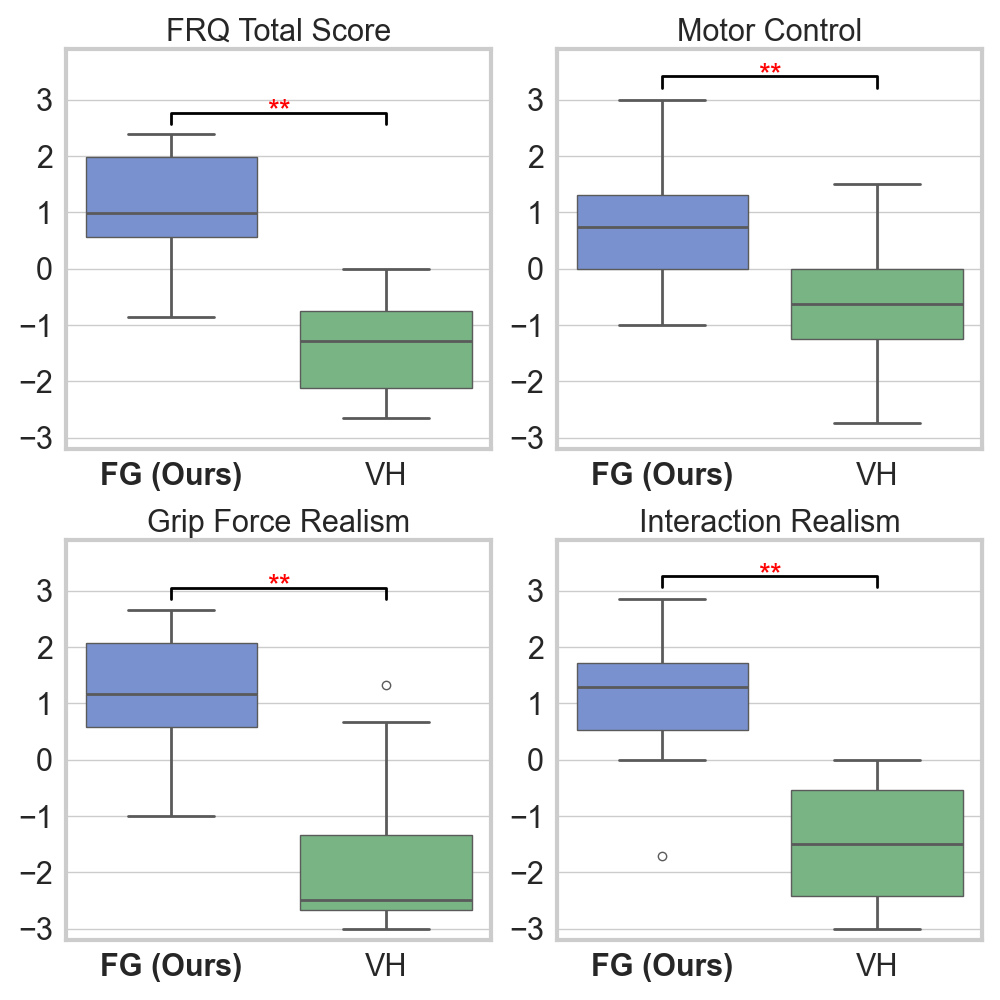}
    \vspace{-0.7cm}
    \caption{FRQ scores for the Can Squeeze task, using a 7-point Likert scale ranging from -3 (strongly disagree) to +3 (strongly agree). Brackets indicate statistical significances ($*$ : $p$ $<$ .05, $**$ : $p$ $<$ .01).}
    \Description{Box plots showing FRQ scores for ForceGrip and VR-HandNet across four metrics: Total Score, Motor Control, Grip Force Realism, and Interaction Realism. ForceGrip shows consistently higher median scores with narrower variability. Statistically significant differences are marked with asterisks.}
    \label{fig:can_squeeze_result}
    \vspace{-0.4cm}
\end{figure}
\subsection{E2: Can squeeze --- Results}

As shown in \autoref{fig:can_squeeze_result}, there exist significant differences between FG and VH across all metrics.
There was no statistically significant change in SSQ scores before and after the experiment.

For FG, many participants reported that ``\textit{finely adjusting the force with the trigger}" felt highly intuitive, suggesting faithful translation of user input to grip force. 
Nonetheless, several noted ``\textit{controlling force with a single index trigger felt unnatural}” (P0, P1, P4, P5, P12) and ``\textit{the lack of haptic feedback makes fine control more challenging}” (P3, P10, P11, P13). Such issues reflect the broader hardware limitations of consumer VR controllers rather than the ForceGrip itself.

For VH, none of the participants provided positive remarks on force control. Instead, they consistently criticized the ``\textit{excessively high grip force regardless of small trigger inputs}” (P1, P2, P3, P4, P5, P6, P8, P15, P19, P20) and described ``\textit{difficult or impossible fine-force adjustments}” (P7, P11, P13, P16).  These reports align with quantitative findings that the method lacks a reliable mechanism to scale force in response to varying user inputs.

\begin{figure}[h!]
    \centering
    \includegraphics[width=1\linewidth]{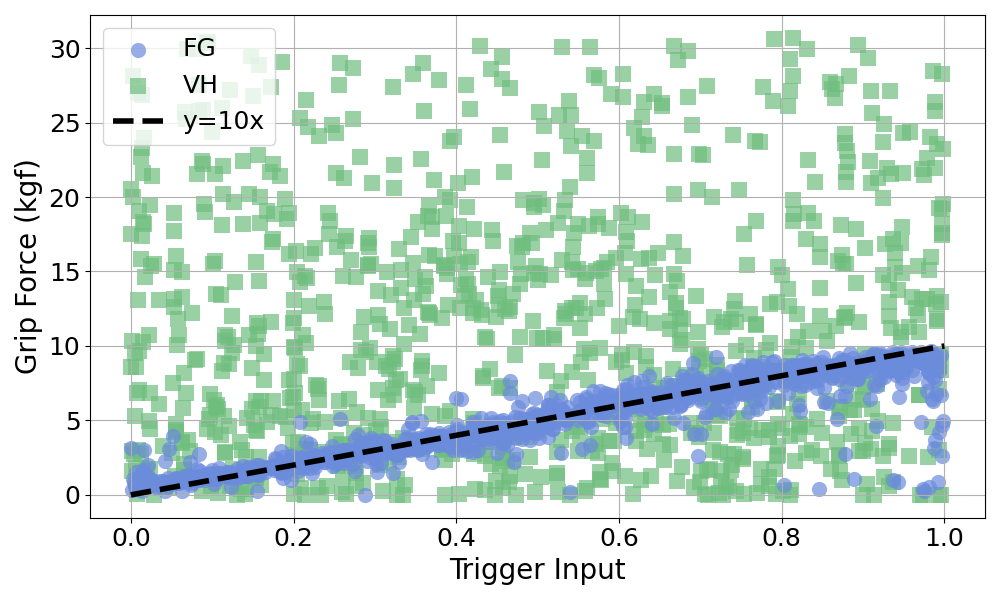}
    \vspace{-0.7cm}
    \caption{Correlation between trigger input ($x$) and grip force ($y$) for FG and VH, with the reference line ($y=10x$). Outliers (top 5\%) were excluded, and the remaining data was duplicated and resampled to 1,000 points, ensuring an even distribution of trigger values. For FG, points below the reference line mostly occurred when the object slipped from the agent's grasp.}
    \label{fig:can_squeeze_trigger_correlation}
    \vspace{-0.2cm}
\end{figure}
To visualize how closely each system reflects user intentions, we plotted the trigger input against the measured grip force (\autoref{fig:can_squeeze_trigger_correlation}). FG exhibited a clear linear correlation ($r = 0.884$). In contrast, VH showed negligible correlation ($r = 0.073$), consistently producing excessive grip forces irrespective of trigger amplitude. This result contradicts the 1.2 kgf maximum force claimed by~\cite{han2023vr} and supports reports that VH appears to lack grip force control.

\section{Conclusion and Future Improvements}
We have introduced \emph{ForceGrip}, a novel approach for delivering realistic hand manipulation in VR controller interfaces through faithful grip force control. Unlike many prior systems that rely on reference datasets, ForceGrip leverages automatically generated training scenarios.
Our curriculum learning framework further addresses the challenges posed by complex, dynamic interactions, enabling robust convergence and state-of-the-art performance across diverse virtual manipulation tasks.

One significant limitation of our method is its reliance on a one-dimensional control signal in the range of [0, 1], chosen for broad compatibility. While this simplifies user input, it limits complex actions such as in-hand manipulation or tool-specific adjustments (e.g., scissors, chopsticks), which would require a richer input structure.
Improving collision and friction modeling through soft-body simulations could produce more stable and lifelike contact dynamics, particularly for slippery or deformable objects.

\begin{acks}
This work was supported by the \grantsponsor{GSNRF}{National Research Foundation of Korea}{https://www.nrf.re.kr/eng/main} under Grant No.~\grantnum{GSNRF}{RS-2025-00518643} (70\%) and \grantnum{GSNRF}{RS-2025-00564137} (10\%).
It was also supported by the \grantsponsor{GSIITP}{Institute of Information \& Communications Technology Planning \& Evaluation (IITP)}{https://www.iitp.kr/en/main.it} grant funded by the Korea government (MSIT) under Grant No.~\grantnum{GSIITP}{IITP-2025-RS-2020-II201819} (10\%) and \grantnum{GSIITP}{RS-2020-II200861} (10\%).
\end{acks}

% Bibliography
\bibliographystyle{ACM-Reference-Format}
\bibliography{bibliography}

\end{document}